%% file: main.tex
\documentclass{article}
\usepackage{ijcai18}

\usepackage{times}
\usepackage{xcolor}
\usepackage{soul}
\usepackage[utf8]{inputenc}
\usepackage[small]{caption}
\usepackage{hyperref}
\usepackage{url}
\usepackage{enumerate}
\usepackage{booktabs}
\usepackage{multirow}
\usepackage{amssymb}
\usepackage{amsmath}
\usepackage{dsfont}
\usepackage{graphicx}
\usepackage{subfig}
\usepackage{wrapfig}

\title{Few-shot Learning by Exploiting Visual Concepts within CNNs}

\author{
Boyang Deng$^1$\thanks{Work done during visiting Johns Hopkins University}, 
Qing Liu$^2$, 
Siyuan Qiao$^2$, 
Alan Yuille$^2$
\\ 
$^1$ Beihang University\\
$^2$ Johns Hopkins University\\
billydeng@buaa.edu.cn,
\{qingliu, siyuan.qiao, alan.yuille\}@jhu.edu
}
\begin{document}

\maketitle

%%%%%%%%%% ABSTRACT
\input{abstract.tex}

%%%%%%%%%% BODY TEXT
\input{section1.tex} % Introduction

\input{section2.tex} % Related Work

\input{section3.tex} % Background: VC

\input{section4.tex} % Few-Shot Learning from VCs

\input{section5.tex} % Evaluation

\input{section6.tex} % Conclusion

\bibliography{references}
\bibliographystyle{named}

\end{document}

%% file: abstract.tex
% !TEX root = main.tex

\begin{abstract}
Convolutional neural networks (CNNs) are one of the driving forces for the advancement of computer vision. Despite their promising performances on many tasks, CNNs still face major obstacles on the road to achieving ideal machine intelligence. One is that CNNs are complex and hard to interpret. Another is that standard CNNs require large amounts of annotated data, which is sometimes hard to obtain, and it is desirable to learn to recognize objects from few examples. In this work, we address these limitations of CNNs by developing novel, flexible, and interpretable models for few-shot learning. Our models are based on the idea of encoding objects in terms of visual concepts (VCs), which are interpretable visual cues represented by the feature vectors within CNNs. We first adapt the learning of VCs to the few-shot setting, and then uncover two key properties of feature encoding using VCs, which we call {\it category sensitivity} and {\it spatial pattern}. Motivated by these properties, we present two intuitive models for the problem of few-shot learning. Experiments show that our models achieve competitive performances, while being more flexible and interpretable than alternative state-of-the-art few-shot learning methods. We conclude that using VCs helps expose the natural capability of CNNs for few-shot learning.
\end{abstract}

%% file: section1.tex
% !TEX root = main.tex

\section{Introduction}
\label{sec:introduction}

%% Background (problem): The black-box property of deep neural network whilst its triumph
After their debut~\cite{lecun1998gradient}, Convolutional Neural Networks (CNNs) have played an ever increasing role in computer vision,
% particularly after their triumph~\cite{krizhevsky2012imagenet} on the ImageNet challenge~\cite{deng2009imagenet}. 
Some researchers have even claimed that CNNs have surpassed human-level performance~\cite{he2015delving}, although other work suggests otherwise~\cite{zhu17object}. Recent studies also show that CNNs are vulnerable to adversarial attacks~\cite{goodfellow2014explaining}. Nevertheless, the successes of CNNs have inspired the computer vision community to develop increasingly sophisticated models~\cite{szegedy2017inception}. 

Despite the impressive achievements of CNNs, we have limited insights into why CNNs are effective. The ever-increasing depth and complicated structures of CNNs make them difficult to interpret while the non-linear nature of CNNs makes it hard to perform theoretical analysis. In addition, CNNs require large annotated datasets which is problematic for many real world applications. We argue that the ability to learn from a few examples, or few-shot learning, is a characteristic of human intelligence and is strongly desirable for an ideal machine learning system.

\begin{figure}[t]
    \begin{center}
        \subfloat[VC\_139\newline(``Sofa Cushion'')]{\includegraphics[width=0.3\linewidth]{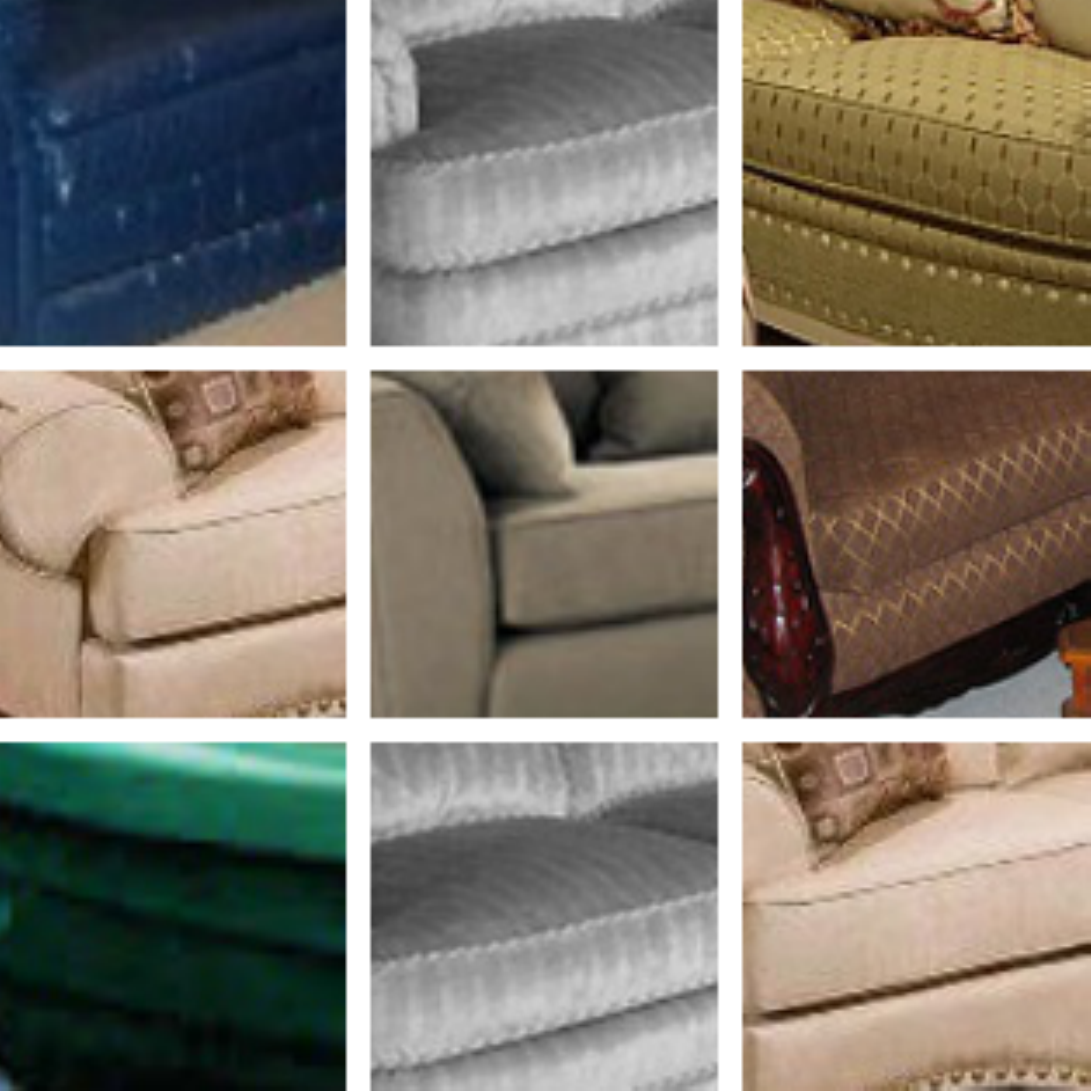}\label{fig:vc_0}}
        \hspace{0.23cm}
        \subfloat[VC\_189\newline(``Side Windows'')]{\includegraphics[width=0.3\linewidth]{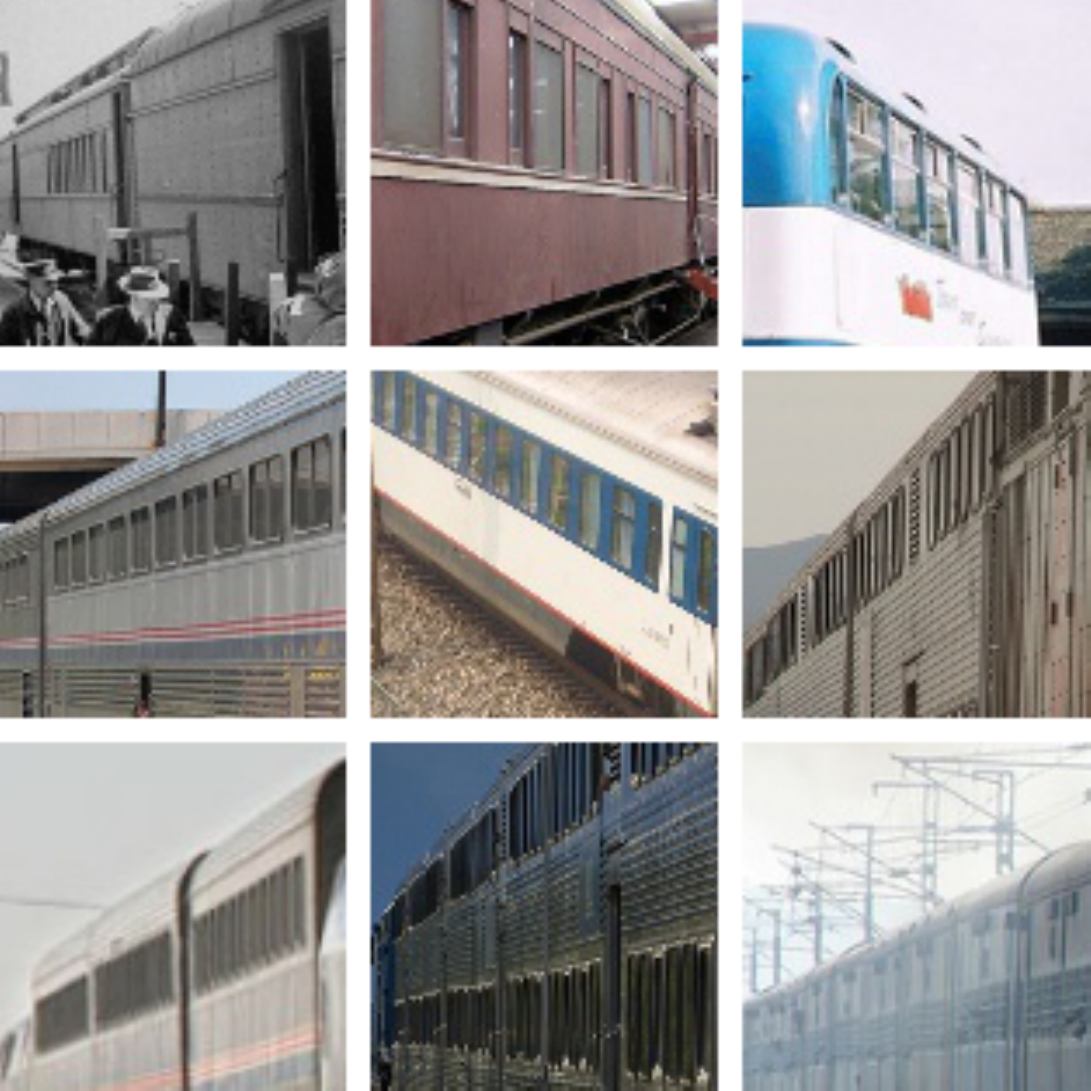}\label{fig:vc_1}}
        \hspace{0.23cm}
        \subfloat[VC\_174\newline(``Bicycle Wheel'')]{\includegraphics[width=0.3\linewidth]{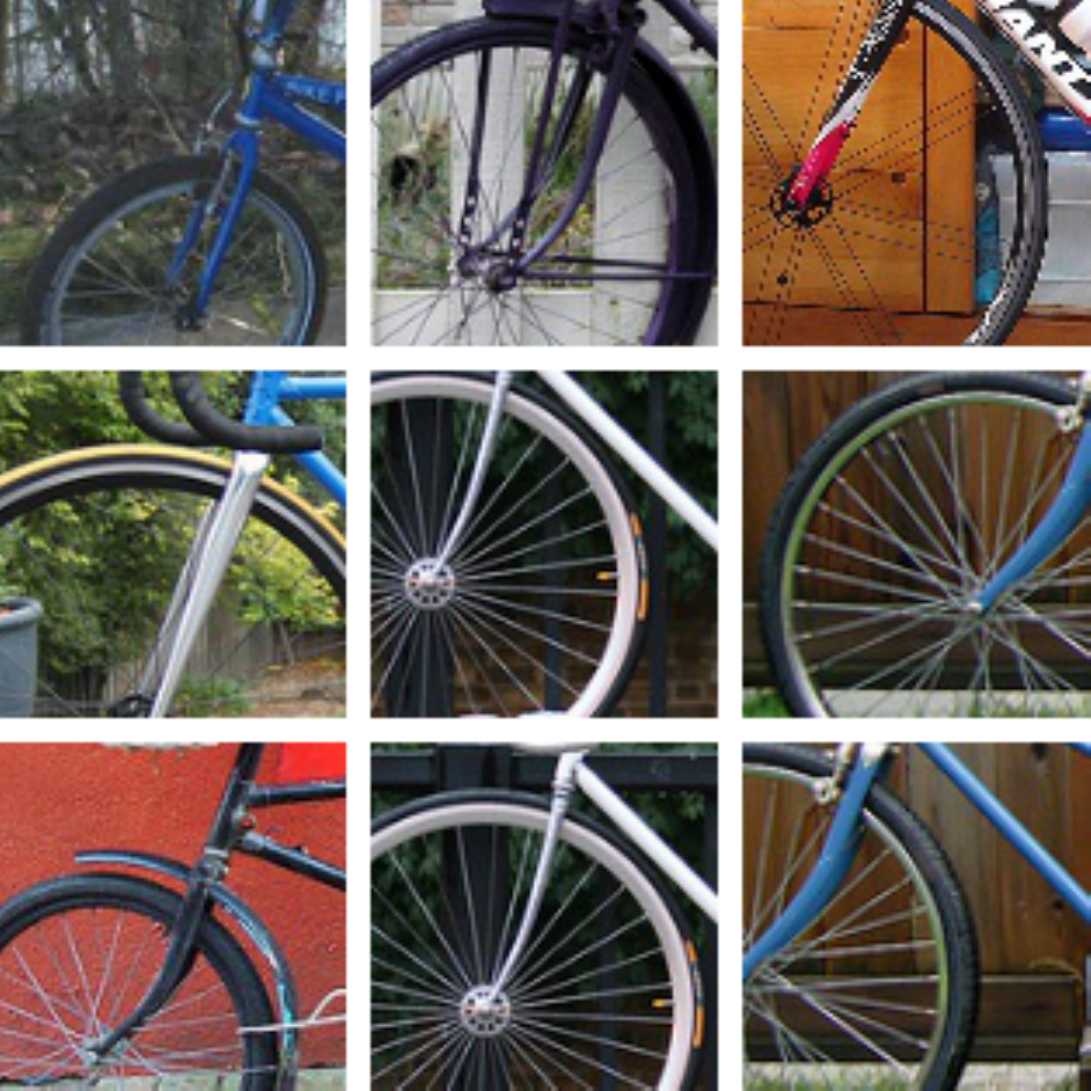}\label{fig:vc_2}}
    \end{center}
    \vspace{-10pt}
    \caption{Visualizations of VCs. Each group consists of patches from original images closest to a VC. In general, these patches roughly correspond to semantic parts of objects, e.g., the cushion of a sofa (a), the side windows of trains (b) and the wheels of bicycles (c). All VCs are referred to by their indices (e.g., VC\_139). We stress that VCs are learned in an unsupervised manner and terms like``sofa cushion'' are inferred by observing the closest image patches and are used to describe them informally.}
\label{fig:vc_vis}
\end{figure}

%% Core Technique
The goal of this paper is to develop an approach to interpretable and flexible few-shot learning which builds on the successes of CNNs. We start from the intuition that objects can be represented in terms of spatial patterns of parts which implies that new objects can be learned from a few examples if they are built from parts that are already known, or which can be learned from a few examples. We recall that previous researchers have argued that object parts are represented by the convolutional layers of CNNs~\cite{zhou2014object,mahendran2015understanding} provided the CNNs are trained for object detection. More specifically, we will build on recent work~\cite{wang2015unsupervised} which learns a dictionary of \textbf{Visual Concepts} (VCs) from CNNs representing object parts, see Figure~\ref{fig:vc_vis}. Their original work proves that these VCs can be combined to detect semantic parts. More recently, it has also been shown that VCs can be used to represent objects using {\bf VC-Encoding} (where objects are represented by binary codes of VCs). 

But it is not obvious that VCs, as described in ~\cite{wang2017detecting}, can be applied to few-shot learning. Firstly, these VCs were learned independently for each object category (e.g., for cars or for airplanes) using deep network features from CNNs which had already been trained on these categories. Secondly, the VCs were learned using large numbers of examples of the object category, ranging from hundreds to thousands. By contrast, for few-shot learning we have to learn the VCs from a much smaller number of examples (by an order of magnitude or more). Moreover, we can only use deep network features which were trained on datasets not including those object categories that we hope to learn within few shots. This means that although we will extract VCs using very similar algorithms to those in~\cite{wang2015unsupervised} our motivation and problem domain are very different. To summarize, in this paper we use VCs to learn models of new object categories from existing models of other categories, while~\cite{wang2015unsupervised} uses VCs to help understand CNNs and to perform unsupervised part detection. Also, unlike \cite{wang2015unsupervised}, we use VC-Encoding.

In Section~\ref{sec:vc}, we will review VCs in detail. Briefly speaking, VCs are extracted by clustering intermediate-level features of CNNs, e.g., features produced by the Pool-4 layer of VGG16~\cite{simonyan2014very}. Serving as the cluster centers in feature space, VCs divide intermediate-level features into a discrete dictionary. We show that VCs can be learned in the few-shot learning setting and they have two desirable properties when used for image encoding, which we call \textit{category sensitivity} and \textit{spatial pattern}.

%% Our findings on VC-Encoding
% More specifically, our method starts by learning a dictionary of visual concepts from a few examples of novel object categories using features from a deep network trained on a different set of objects. Unlike~\cite{wang2015unsupervised} these visual concepts are shared between the novel object categories (there is not enough data to learn visual concepts for each novel object category separately). The novel objects can then be represented in terms of spatial patterns of the binary encoding of the objects, which we call VC-Encoding. We make   three preliminary findings. Firstly, some VCs are closely related to specific object categories, \textit{i.e.}, they mainly occur in images from a specific object category. Secondly, VC-Encodings result in regular spatial patterns in images as shown by the heat maps in Figure~\ref{fig:ppt2}. Thirdly, that the VCs are training cost effective in the sense that several images are sufficient for extracting VCs for VC-Encoding. These three findings encouraged us to proceed to Few-Shot Learning. 

More specifically, we develop an approach to few-shot learning which is flexible and interpretable. We learn a dictionary of VCs as described above which enables us to represent novel objects by their VC-Encoding. Then we propose two intuitive models: (i) nearest neighbor and (ii) a factorizable likelihood model based on the VC-Encoding. The nearest neighbor model uses a similarity measure to capture the difference between two VC-Encodings. The factorizable likelihood model learns a likelihood function of the VC-Encoding which, by assuming spatial independence, can be learned from few examples. We emphasize that both these models are very flexible, in the sense that they can be applied directly to any few-shot learning scenarios. This differs from other approaches which are trained specifically for scenarios such as 5-way 5-shot (where there are 5 object categories with 5 examples of each). This flexibility is attractive for real world applications where the numbers of new object categories, and the number of examples of each category, will be variable. Despite their simplicity, these models achieve comparable results to the state-of-the-art few-shot learning methods, such as \emph{learning a metric} and \emph{learning to learn}. From a deeper perspective, our results show that CNNs have the potential for few-shot learning on novel categories but to achieve this potential required studies of the internal structures of CNNs to re-express them in simpler and more interpretable terms.

%% Few-Shot Learning

%% Evaluations of our model

%% Our contributions
Overall, our major contributions are two-fold:

\begin{enumerate}[(1)]
\item We show that VCs can be learned in the few-shot setting using CNNs trained on other object categories. By encoding images using VCs, we observe two desirable properties, i.e., category sensitivity and spatial pattern.
\item Based on these properties, we present two simple, interpretable, and flexible models for few-shot learning. These models yield competitive results compared to the state-of-the-art methods on specific few-shot learning tasks and can also be applied directly, without additional training, to other few-shot scenarios.
\end{enumerate}

%% file: section2.tex
% !TEX root = iclr_main.tex

\section{Related Work}
\label{sec:relwork}
Our work on few-shot learning is motivated by and builds on attempts to understand the internal representations of neural networks. Therefore, we review here the previous literature on these topics.

\subsection{Neural Network Internal Representations}
Recently, there have been numerous studies aimed at understanding the behavior of neural networks and, in particular, to uncover the internal
representations within CNNs. Some try to visualize internal representations by sampling~\cite{zeiler2014visualizing}, generating~\cite{simonyan2013deep} or by backpropagating~\cite{mahendran2015understanding} images in order to maximize the activations of the hidden units. A particularly relevant work by~\cite{zhou2014object} shows that object and object parts detectors emerge in CNNs.
Conversely, other works investigate the discriminative power of the hidden features of CNNs by assessing them on specific problems~\cite{sharif2014cnn,netdissect2017,yosinski2014transferable}. The overall findings suggest that deep networks have internal representations of object parts. The most relevant work to our paper is the study of VCs which discovered mid-level visual cues in the internal features of CNNs and showed relationships between these visual cues and semantic parts~\cite{wang2015unsupervised,wang2017detecting}. This work is described in detail in Section \ref{sec:vc}

\subsection{Few-Shot Learning}

There have been growing attempts to perform few-shot learning motivated by attempts to mimic human abilities and to avoid some of the limitations of conventional data-demanding learning. An early attempt was made building on probabilistic program induction~\cite{lake2015human}. More recent efforts at few-shot learning can be broadly categorized into two classes. The first is to design methods to embed the inputs into a feature space friendly to few-shot settings~\cite{koch2015siamese,vinyals2016matching}. Their goal is to find a good similarity measure (e.g., using Siamese networks) that can be applied rapidly to novel categories. The second is meta-learning which efficiently trains an ordinary model with the budget of few examples~\cite{ravi2016optimization,finn2017model}. An alternative approach by~\cite{qiao2017few} performs few-shot learning by estimating parameters of the prediction layer using regression from previously learned objects. We emphasize that the approach in our paper differs from these works, many of which are tailored for a few specific few-shot learning scenarios (i.e., test and train conditions must match), while our methods are simple and flexible, so they work both in normal and almost all few-shot settings.

%% file: section3.tex
% !TEX root = main.tex

\section{Background: Visual Concepts}
\label{sec:vc}
\begin{figure}[t]
    \begin{center}
    \includegraphics[width=0.9\linewidth]{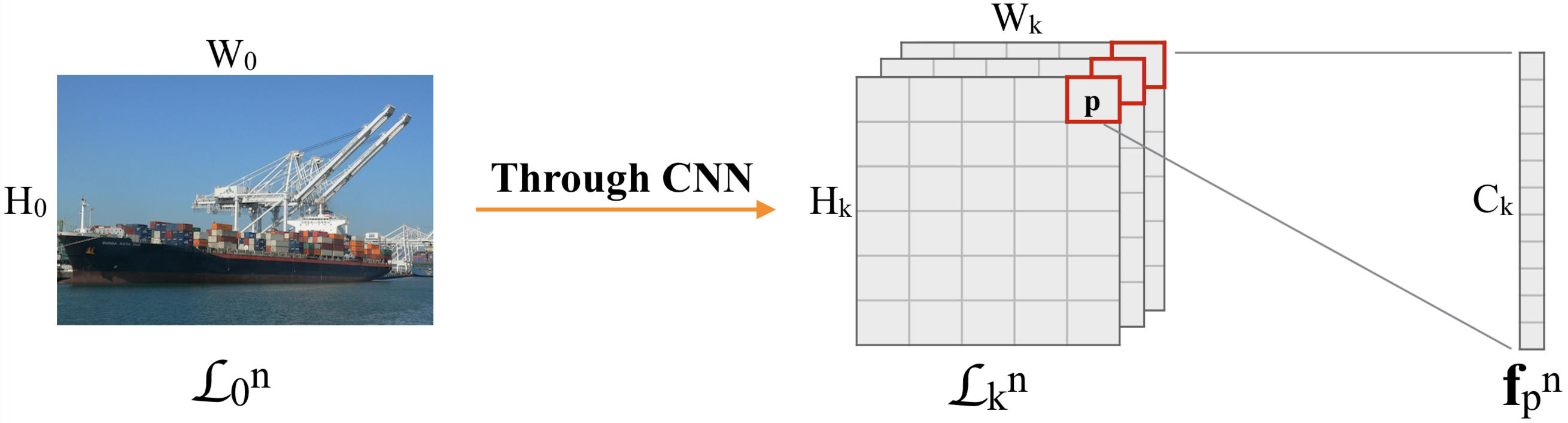}
    \end{center}
    \vspace{-10pt}
    \caption{Key terms in the VC formalization. On the left is the $n$-th input image, defined on image lattice $\mathcal{L}_0^n$, with height $H_0$ and width $W_0$. In the middle is the lattice at the $k$th layer of the CNN for the $n$-th image, noted by $\mathcal{L}_k^n$, with height $H_k$ and width $W_k$. On the right is a feature vector at position $p$ in $\mathcal{L}_k^n$, noted by $\mathbf{f}_p^n$, with dimensionality $C_k$.}
    \label{fig:vc_form}
\end{figure}
In~\cite{wang2015unsupervised}, VCs were discovered as internal representations within deep networks
which roughly correspond to mid-level semantic visual cues. These VCs play a core role in our work on understanding properties of CNNs and developing our interpretable few-shot learning models. In this section, we review how VCs were defined and learned in~\cite{wang2015unsupervised}. We will describe later in Section \ref{subsec:fewshotvcs} how we modify VCs for few-shot learning.

We first summarize the  formalization of VCs, which are illustrated in
Figure~\ref{fig:vc_form}. CNNs contain a hierarchy of lattices $\mathcal{L}_l$, where $l
\in \{0, 1,\ldots\}$ stands for the layer of the lattice. In particular, the input image is 
defined over the lattice $\mathcal{L}_0$ and the lattice on which we derive VCs is specified by $\mathcal{L}_k$. 
We denote the spatial mappings from $\mathcal{L}_0$ to $\mathcal{L}_k$ by $\pi_{0 \mapsto
k}$ and from $\mathcal{L}_k$ to $\mathcal{L}_0$ by $\pi_{k \mapsto 0}$. Then we define
$\mathcal{F}_k^n =\{\mathbf{f}_p^n : p \in \mathcal{L}_k^n\}$ as the feature vector set at
$\mathcal{L}_k^n$ of the $n$-th image where $p$ refers to a 2D position in the lattice $\mathcal{L}_k^n$.
These feature vectors are computed by $\mathbf{f}_p^n = f(\mathbf{I}_{A(p')}^n)$, where the
function $f$ is specified by the neural network and $\mathbf{I}_{A(p')}^n$ is a subregion
of the $n$-th input image $\mathbf{I}^n$, centered on a point $p'=\pi_{k \mapsto 0}(p)$ on
$\mathcal{L}_0^n$.  
In other words, the responses of all the channels in position $p$ constitute the feature
vector $\mathbf{f}_p^n$. 
Then we have $\mathcal{F}_k = \cup_n \mathcal{F}_k^n$ from all images of interest. 
Note that by collecting feature vectors into $\mathcal{F}_k$, all spatial and image identity information is removed. Since layer $k$ is usually pre-selected for different network architectures for VCs applications (e.g., \cite{wang2015unsupervised} typically studied layer $k=4$ for VGG16 net in their work), the subscript will be omitted in the remainder of the paper for simplicity.

Now we describe how VCs are extracted. The approach assumes that the VCs are represented by a population code of the CNN feature vectors. They are extracted using an unsupervised clustering algorithm. Since we first normalize the feature vectors into unit length, instead of using K-means as proposed in~\cite{wang2015unsupervised}, we assume that the feature vectors
are generated by a  mixture of von Mises-Fisher distributions (vMFM) and
learn this mixture by the EM algorithm~\cite{banerjee2005clustering}. The goal is to maximize the likelihood function
\begin{equation}
\begin{small}
P(\mathcal{F}|\alpha, \mu, \kappa) = \Pi_{m=1}^M\Sigma_{v=1}^V\alpha_{m,v}V_d(\mathbf{f}_m|\mu_v, \kappa_v),
\end{small}
\end{equation}
\noindent where $M$ is the total feature vector count (i.e., the number of all feature vectors collected from all images of interest) and $V$ is the predefined VC (cluster) number. $V_d(\cdot)$ is the density function of the vMF distribution. $\mathbf{f}_m$ denotes each feature vector we get from the intermediate layer of a CNN (without image identity or spatial information). $\alpha$, $\mu$, and $\kappa$ are vMFM parameters and represent the mixing proportion, mean direction, and concentration values respectively.

We define the set of VCs by $\mathcal{V} =\{\mu_v : v = 1, \dots, V\}$ (i.e., by the mean directions of the learned vMFM distribution). Alternatively, since the $\{\mu_v\}$ have the same dimensionality as the $\{\mathbf{f}_m\}$, we denote a specific VC center by $\mathbf{f}_v = \mu_v$.

To help understand the VCs, we compute the cosine distances from the original feature vectors to the VCs as follows:
\begin{equation}
\begin{small}
    d_{p,v}^n=1 - \frac{\mathbf{f}_p^n \cdot \mathbf{f}_v}{\left\lVert  \mathbf{f}_p^n \right\rVert _2\left\lVert \mathbf{f}_v \right\rVert_2 },
\label{equ:distance}
\end{small}
\end{equation}
\noindent where $d_{p,v}^n$ denotes the distance between feature vector $\mathbf{f}_p^n$ and the VC $v$ in the $n$-th image at position $p$, and we call them \textbf{VC distances}.
We select those feature vectors with the smallest distances to each VC and trace them
back to the original input image using $\pi_{k \mapsto 0}$. This yields ``visualization patches'' of VCs, shown in Figure~\ref{fig:vc_vis}. We observe 
that these patches roughly correspond to the semantic parts of objects, which justifies our  
assertion that VCs are semantic visual cues. 

% In the original paper~\cite{wang2015unsupervised} VCs were evaluated as semantic part detectors. Later work~\cite{wang2017detecting} showed that VCs could be combined by voting to detect semantic parts and performed well even if the parts were partially occluded. 

In the previous studies of VCs \cite{wang2015unsupervised,wang2017detecting}, the CNNs that were used to generate feature vectors were trained for a large scale object classification task that included the object categories of interest. Moreover, they extracted VCs using hundreds of images within a specific category of object, which resulted to category specific visual cues that were useful for interpreting CNN behaviors and building novel models for semantic part detections. In more recent work (in preparation) VCs were used to encode semantic parts and objects using \emph{VC-Encoding} that could be applied to detection tasks in the presence of occlusion. VC-Encoding is described in the next section. We emphasize that none of this prior work on VCs addressed few-shot learning and only addressed situations where there were many training examples of the object categories. 

%% file: section4.tex
% !TEX root = iclr_main.tex

\section{Few-Shot Learning from VCs}
\label{sec:classification}

This section describes the technical ideas of our paper. In Section~\ref{subsec:fewshotvcs}, we introduce how we learn VCs in the few-shot setting. In Section~\ref{subsec:properties}, we introduce VC-Encoding and show its two desirable properties for few-shot classification tasks. 
%Then we discuss three important properties which motivate the use of VCs for few-shot learning. 
%We note that VC-Encoding was developed for another research project (in preparation) but the three properties are contributions of this paper.
Then in Section~\ref{subsec:nn} and Section~\ref{subsec:factor}, we propose two simple and interpretable VC-Encoding models for few-shot learning. 

\subsection{Few-shot VCs}
\label{subsec:fewshotvcs}
It is not obvious that VCs can be applied to few-shot learning tasks where only few examples are available for each novel category. It is not possible to train the CNNs on all the object categories (as was done in~\cite{wang2015unsupervised}) and also there may not be enough data to get good VC clusters. Hence we modify the way VCs are learned: we learn VCs from small number of examples of novel object categories using features from CNNs trained on other object categories. This is similar to how metric-learning and meta-learning are trained on large datasets which do not include the novel categories and ensures that the CNN used for feature extraction has never seen the categories on which we will perform the few-shot task. To extract VCs for the novel categories which only have few examples each, we pool feature vectors from the different categories together and perform the clustering algorithm on all of them. This gives us a little more data and encourages VC sharing between different categories, which also makes it easier to apply our VC models to multiple novel categories.  

By the two modifications described above, we obtain \textbf{few-shot VCs}, i.e., VCs that are suitable for few-shot learning. This is critical for our application and differentiates this work from previous studies of VCs (e.g., \cite{wang2015unsupervised}). Surprisingly, we find that we only need a few images (e.g., five images per category) to extract high quality VCs (see visualizations in Figure \ref{fig:ppt1}) which, when used for VC-Encoding, possess similar desirable properties as the traditional VCs and hence are suitable for few-shot object classification task. 

\subsection{VC-Encoding}
\label{subsec:properties}
\begin{figure}[t]
    \begin{center}
        \subfloat[Category Sensitivity]{\includegraphics[width=0.35\linewidth]{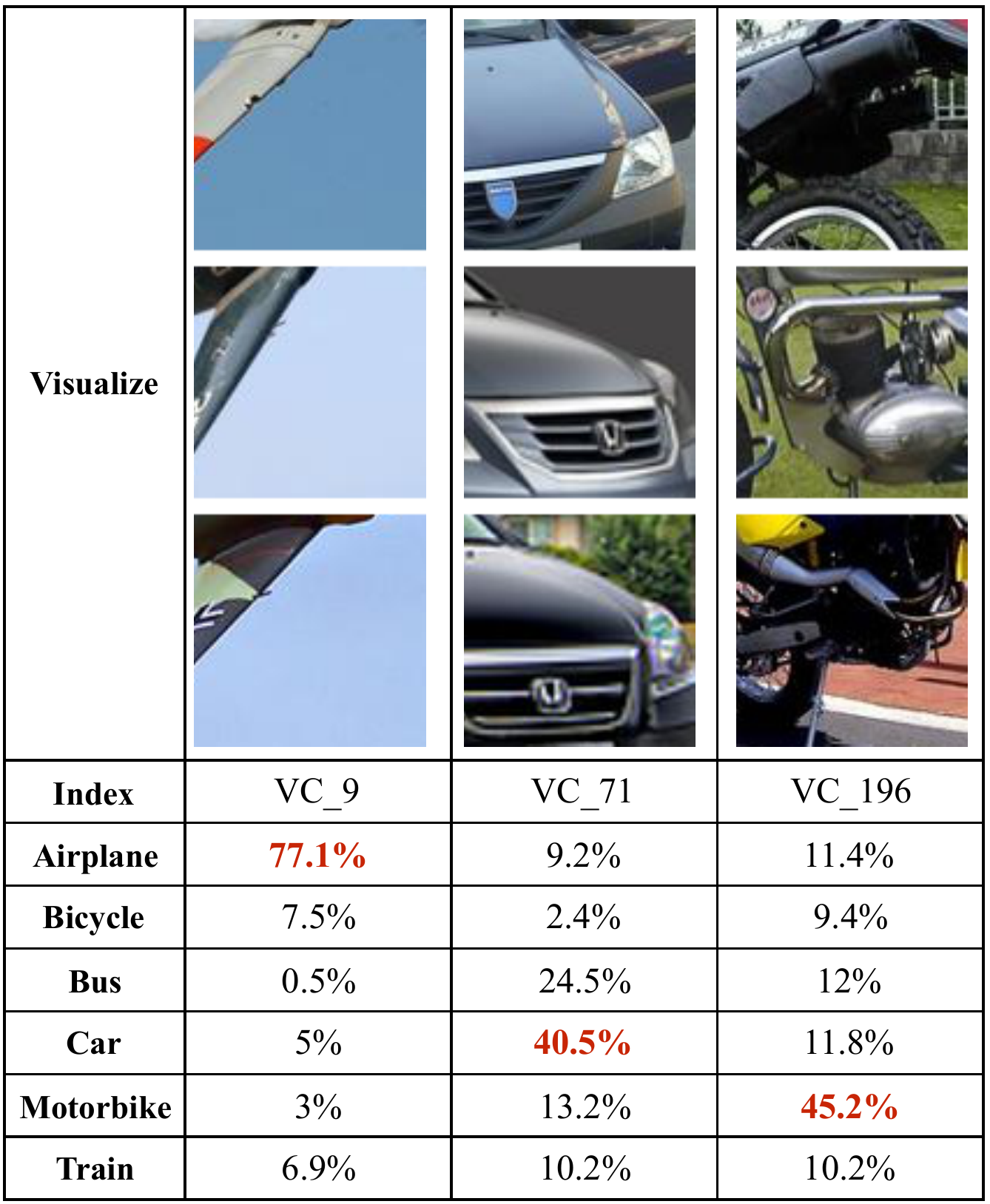}\label{fig:ppt1}}
        \hspace{0.2cm}
        \subfloat[Spatial Pattern]{\includegraphics[width=0.53\linewidth]{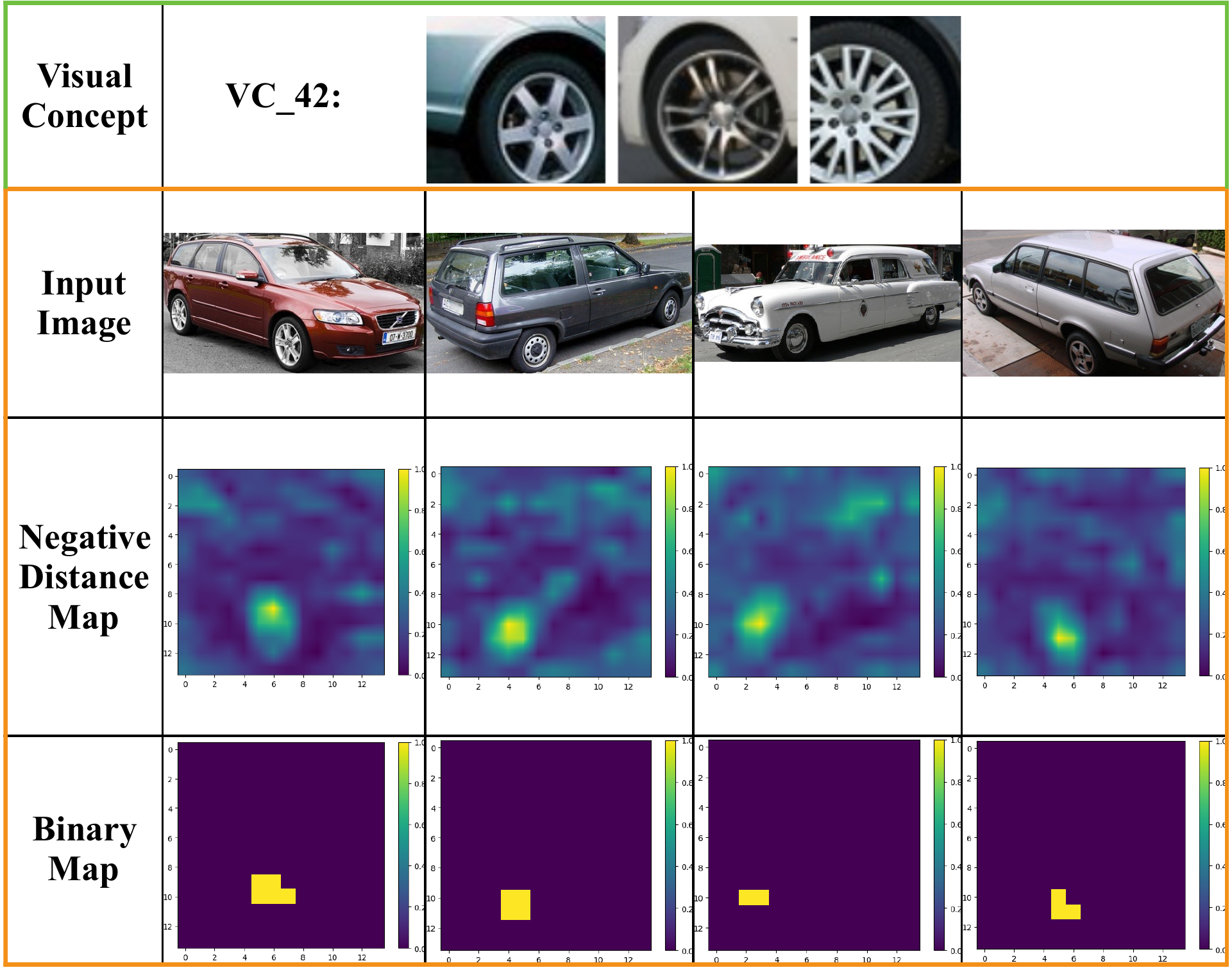}\label{fig:ppt2}}
    \end{center}
    \vspace{-10pt}
    \caption{Properties of VCs. In (a), we illustrate three VCs by their closest patches and their occurrence distributions over $6$ object categories out of the $12$ in PASCAL3D+ showing category sensitivity of VC-Encoding. In (b), we visualize the closest patches to VC\_42 in the top green box and randomly select $4$ images of cars with their negative distance maps and binary maps w.r.t. VC\_42 plotted in the bottom orange box. The negative distance map is given by $-d_{p,42}$ and is scaled to $(0,1)$. The binary map is drawn based on $b_{p,42}$. See Section \ref{subsec:properties} for more details.}
\label{fig:ppts}
\end{figure}

We assume that objects can be decomposed into semantic parts. From the perspective of
VCs, this means that most $\{\mathbf{f}_p\}$ should be assigned to \emph{a single VC}. This requires
specifying an explicit relationships between the $\{\mathbf{f}_p\}$ and the VCs. A natural choice is to
compute the distances $d_{p,v}$ between the $\mathbf{f}_p$ and the $v$-th VC and threshold it to produce
a binary value $b_{p,v}$ (i.e., $b_{p,v}=1$ if $d_{p,v} < T$). We refer to $\mathcal{B} = \{b_{p,v}: p \in \mathcal{L}, v=1, \dots, V\}$ as the \textbf{VC-Encoding}. Note the image index $n$ is omitted here since the operations are identical for all images of interest. We use two criteria to specify a good encoding, $coverage$ and 
$firerate$, defined as following:
\begin{small}
\begin{eqnarray}
    coverage &=& \frac{\sum_{p\in \mathcal{L}} \max_v b_{p,v}}{\left\lvert \mathcal{L} \right\rvert}, \\
    firerate &=& \frac{\sum_{p\in \mathcal{L}} \sum_v b_{p,v}}{\left\lvert  \mathcal{L} \right\rvert}.
\label{equ:covfir}
\end{eqnarray}
\end{small}
The choice of the encoding threshold $T$ is a trade-off between requiring sufficient coverage and a firing rate that is close to one. In practice, we choose $T$ for each testing trial by a grid-search with step size $0.001$ which outputs the smallest threshold ensuring that the average $coverage>=0.8$ for all few-shot training images. This yields the final VC-Encoding $\mathcal{B}$ used in our models, with the following desirable properties:
%$ = \{b_{p,v} = \mathds{1}(d_{p,v} < T): p \in \mathcal{L}, v=1, \dots, V\}$.

\textbf{Category Sensitivity} Despite the fact that the VCs are learned from a mix of images with different category labels, the first insight is that many VCs
tend to fire ($b_{\cdot, v}=1$) intensively for one or a few specific object categories. In Figure~\ref{fig:ppt1}, we
calculate the occurrence distributions of several VCs for $6$ object categories out of the $12$ in
PASCAL3D+~\cite{xiang2014beyond}. In each column that represents a specific VC, the occurrence frequencies tend to be high for one or two object categories and low for the others. This suggests that VC identities can provide useful informations for object classification.
Moreover, the corresponding visualized patches on the top of Figure~\ref{fig:ppt1} support our understanding that VCs have this category sensitivity because they capture the semantic parts that are specific for object categories.

\textbf{Spatial Pattern} The spatial pattern of VC firings is also indicative of the object category.
Although spatial information is ignored during feature clustering, the learned VCs give binary maps that contain regular
spatial patterns for images of the same category with relatively similar viewpoints (as shown in Figure~\ref{fig:ppt2}). This is consistent with the more general conjecture that the spatial patterns of semantic parts play a vital role in object recognition, and shows again that the VC-Encoding can capture the spatial patterns of the semantic parts to a certain extend.

Next, we design two simple few-shot learning models based on VC-Encoding learned from few examples.
% \textbf{Training Efficiency} After showing that VC-encoding is potentially useful for object classification (the first two properties), we then study how many examples are needed to learn good VCs. This is critical for learning in few-shot settings and differentiates our work with previous applications of VCs the most. We find that we only need a few images in order to extract VCs which, when used for VC-encoding, possess the first two properties and hence are suitable for object classification task. 
% These frequencies are very similar to those when many images are used to train the visual concepts. We observe, for example, that only the car and bus respond to the wheel visual concept. As we will show, these frequencies can be used to build models learnt from few-shots.

\subsection{Nearest Neighbor on Spatial Patterns}
\label{subsec:nn}

First, we propose a simple template matching model which is
similar to traditional nearest neighbor algorithms. The novelty is that we use
a similarity metric between VC-Encodings which is spatially ``fuzzy'' so that it can tolerate small spatial shifts of the parts in images. Formally, the similarity metric takes the following form:
\begin{small}
\begin{equation}
    K(b, b^\prime) = \frac{1}{2}(\frac{\sum_{p,v}b_{p,v}\max_{q,q\in
n(p)}b_{q,v}^\prime}{\sum_{p,v}b_{p,v}} + \frac{\sum_{p,v}b_{p,v}^\prime \max_{q,q\in
n(p)}b_{q,v}}{\sum_{p,v}b_{p,v}^\prime}),
\end{equation}
\end{small}
\begin{wrapfigure}[6]{r}{0.45\linewidth}
    % \begin{center}
    \centering
    \includegraphics[width=\linewidth]{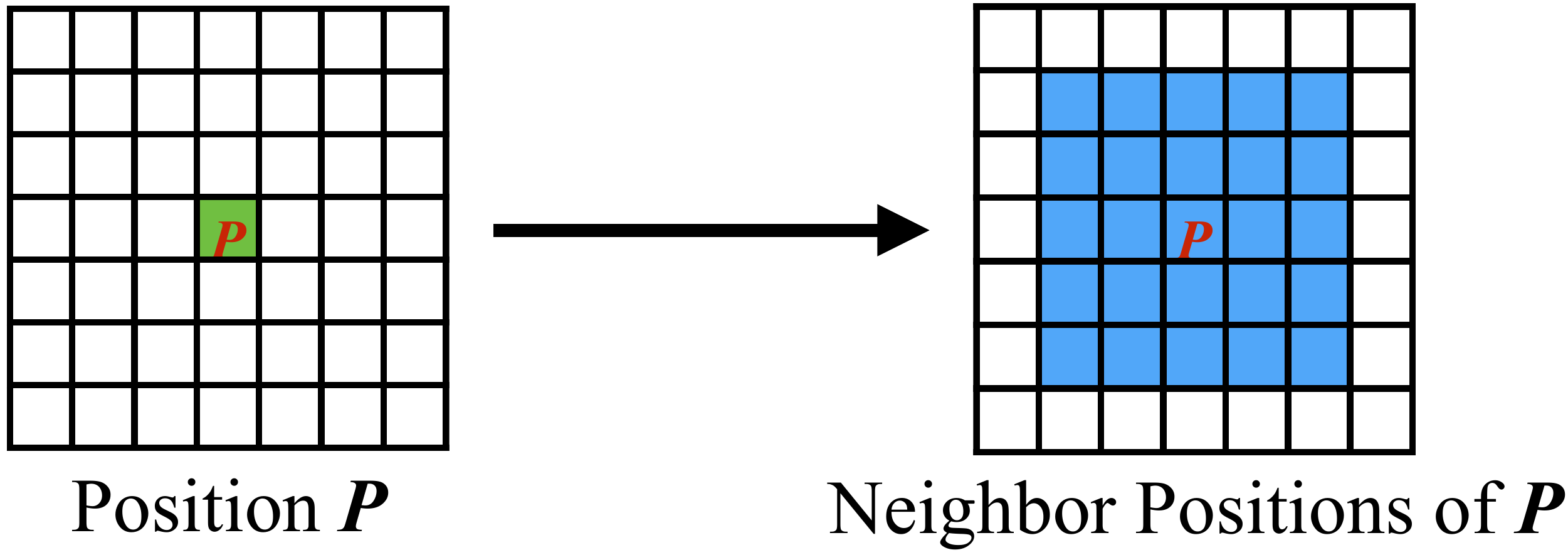}\\
    % \end{center}
    \vspace{-10pt}
    \caption{}
    \label{fig:neighbors}
\end{wrapfigure}
% The green grid on the left is $p$. The blue grids on the right are $n(p)$.
\noindent where $K(b, b^\prime)$ is the similarity between the binary VC-Encodings $b$ and $b^\prime$. $n(p)$ defines the set of neighboring positions of $p$
(as shown in
Figure~\ref{fig:neighbors}). During testing, we classify an image to the category of the training example with the largest similarity.

% One motivation for this method is that sequential convolutional operations carried in a neural network can be considered as embedding input images into a hierarchy of feature spaces. Each convolutional layer can be treated as a different level of decomposition of the inputs since the convolution along with non-linear activation, which take the form $\mathbf{Y}=\sigma(\mathbf{W}\cdot \mathbf{X} + \mathbf{B})$, is composed of matching templates $\mathbf{W}$ and using non-linear function $\sigma$ to filter out patterns based on the threshold $\mathbf{B}$. In light of this interpretation, and the two properties described earlier, it is reasonable that VC-Encoding will yield an explicit semantic decomposition.

\subsection{Factorizable Likelihood Model}
\label{subsec:factor}
Apart from the intuitive nearest neighbor method, we present a second method which
models the likelihood of the VC-Encoding. We observe that we can specify a distribution over the VC-Encoding
 $b_{p,v}$ using a bernoulli distribution with probability $\theta_{p,v}$. Following  Na\"ive
Bayes, we assume all the elements of the VC-Encoding  $b$ are independent (making it possible to learn the distribution from a very small number of examples). Hence we can
express the likelihood of $b$ as following:
\begin{small}
\begin{equation}
    \mathcal{L}(b | \theta) = \prod_{p,v}b_{p,v} \cdot \theta_{p,v} + (1 - b_{p,v}) \cdot (1 - \theta_{p,v}).
\end{equation}
\end{small}
For each object category $y$, we derive a probabilistic distribution $\theta_y$ from the training
examples. Thus the prediction of object category given the VC-Encoding $b$ is given by:
\begin{small}
\begin{equation}
    y_b = \max_y \mathcal{L}(b | \theta_y).
\end{equation}
\end{small}
Note that by doing this, we are in fact implementing a discriminative model obtained from a
generative distribution. We smooth each distribution $\theta_y$ using a Gaussian filter to guard against unlikely events.

%% file: section5.tex
% !TEX root = main.tex

% \section{Evaluations under Few-Shot Settings}
\section{Experiments}
\label{sec:fewshot}
Section~\ref{subsec:fewshotvcs} suggests that a few images may be enough for learning object models when represented by VC-Encodings. Indeed our experiments show that both our two VCs-based few-shot learning models are competitive in performance with alternative methods designed specifically for few-shot learning such as \cite{ravi2016optimization}. In addition, while previous few-shot methods are trained to work in specific few-shot scenarios, such as 5-way classifications, our methods can be applied to a large range of  few-shot scenarios without additional training. The experimental results show that trained CNNs have the potential to recognize novel objects from few examples by exploiting VC-Encoding.
% Few-Shot learning is a very challenging task where humans perform much better than current algorithms. It requires the ability, or efficiency, to learn generalizable knowledge from strictly limited examples, such as a few training images. Nevertheless,

\subsection{Mini-ImageNet}
\label{subsec:miniimagenet}

To assess the capability of our few-shot methods, we first evaluate them on a common few-shot
learning benchmark, namely Mini-ImageNet. The Mini-ImageNet dataset was first proposed by \cite{vinyals2016matching} as a benchmark
for evaluating few-shot learning methods. It selects $100$ categories out of 
$1000$ categories in ImageNet with $600$ examples per category. We use the split proposed by \cite{ravi2016optimization} consisting of
$64$ training categories, $16$ validation categories and $20$ testing categories. In accordance
with the convention for Mini-ImageNet, we perform numerous trials of few-shot learning during testing. In each
trial, we randomly sample $5$ unseen categories from a preserved testing set. Each category is
composed of $5$ training images for the 5-shot setting and $1$ training image for the
1-shot setting. During evaluation, we randomly select $15$ images for each category following~\cite{ravi2016optimization}.

% Mini-ImageNet
\begin{table}\footnotesize
    \begin{center}
    \begin{tabular}{l|cc} 
        \toprule
        \multirow{2}{*}{\textbf{Method}} & \multicolumn{2}{c}{\textbf{5-category}}\\ % &
        % \textbf{10-category} \\
        & 1-shot & 5-shot \\
        \midrule
        \textbf{Baseline-FT}  & $28.86 \pm 0.54\%$ & $49.79 \pm 0.79\%$ \\ 
        \textbf{Baseline-NN} & $41.08 \pm 0.70\%$ & $51.04 \pm 0.65\%$ \\
        \textbf{Pool3-NN} & $43.38 \pm 0.81\%$ & $55.33 \pm 0.75\%$ \\
        \textbf{MatchingNet} & $43.56 \pm 0.84\%$ & $55.31 \pm 0.73\%$ \\
        \textbf{Meta-Learner} & $43.44 \pm 0.77\%$ & $60.60 \pm 0.71\%$ \\
        \textbf{MAML} & \boldmath $48.70 \pm 1.84\%$ & \boldmath $63.11 \pm 0.92\%$ \\
        \midrule\midrule
        \textbf{VC-NN (Ours)} & \boldmath $46.39 \pm 1.09\%$ & $58.84 \pm 1.12\%$ \\
        \textbf{VC-LH (Ours)} & $45.61 \pm 1.14\%$ & \boldmath $63.07 \pm 1.02\%$ \\
        \bottomrule
    \end{tabular}
    \end{center}
    \vspace{-10pt}
    \caption{Average classification accuracies on Mini-ImageNet with $95\%$ confidence
    intervals. Evaluations of Baseline-FT(finetune) and Baseline-NN(nearest neighbor) are
    from~\protect\cite{ravi2016optimization}. Pool3-NN stands for a nearest neighbor method based 
    on raw Pool-3 features from the same VGG-13 as our methods. At the bottom are our nearest neighbor method (VC-NN) and factorizable likelihood method (VC-LH) based on VCs. Marked in bold at the top are the best published results for each scenario. Marked in bold at the bottom are our best results for the corresponding set-up. Note we adopt the results for Matching Network from~\protect\cite{ravi2016optimization}.}
\label{tab:miniimagenet}
\end{table}

\begin{table}[b]\footnotesize
    \begin{center}
    \begin{tabular}{l|cc} 
        \toprule
        \multirow{2}{*}{\textbf{Method}} & \textbf{6-category} & \textbf{8-category} \\ % & \textbf{12-category} \\
        & 3-shot & 4-shot \\ % & 6-shot \\
        \midrule 
        \textbf{Baseline-NN} & \boldmath $46.70 \pm 0.84\%$ & $42.48 \pm 0.74\%$ \\ 
        % $38.49 \pm 0.49\%$ \\
        \textbf{Pool3-NN} & $44.25 \pm 0.73\%$ & \boldmath $43.30 \pm 0.73\%$ \\ 
       % \boldmath $38.77 \pm 0.53\%$ \\
        \midrule\midrule
        \textbf{VC-NN (Ours)} & $50.42 \pm 0.97\%$ & $46.39 \pm
        0.74\%$ \\ % & $40.78 \pm 0.54\%$ \\
        \textbf{VC-LH (Ours)} & \boldmath $52.41 \pm 0.93\%$ & \boldmath $47.37 \pm 0.74\%$ \\
        % & \boldmath $43.42 \pm 0.54\%$ \\
        \bottomrule
    \end{tabular}
    \end{center}
    \vspace{-10pt}
    \caption{Average classification accuracies on Mini-ImageNet with $95\%$ confidence
    intervals under randomly selected few-shot settings. All models used here are the same as the ones used in Table~\ref{tab:miniimagenet}. Our method adapts easily to different number of categories and different number of shots with minimal re-training and consistently outperform baseline methods, while the other state-of-the-art cannot be directly applied to these settings.}
\label{tab:flexible}
\end{table}

For our methods, we train a VGG-13 on the training and validation set with the objective of cross entropy. We preserve $10\%$ images in each category to validate our network. Then, we extract $200$ VCs from the Pool-3 layer for the current testing pool. The reason for choosing Pool-3 features is that a grid in Pool-3 lattices $\mathcal{L}_3$ correspond to a $36\times36$ patch in the original $84\times84$ image, which is a plausible size for a semantic part. For the Gaussian filter used to smooth the factorizable likelihood model, we use $\sigma=1.2$.

As Table~\ref{tab:miniimagenet} illustrates, we compare our methods against two baselines in line with the ones in \cite{ravi2016optimization} (referred to by Baseline-FT and Baseline-NN). To directly examine the impact of the VCs, we also include the result of nearest neighbor matching using raw features from the pool-3 layer (referred to by Pool3-NN). In addition, we present the performances of state-of-the-art few-shot learning methods, including MatchingNet~\cite{vinyals2016matching}, Meta-Learner~\cite{ravi2016optimization}, and MAML~\cite{finn2017model}. The results show that our VCs-based methods compare well with current methods which were specifically designed for few-shot learning. Compared with meta-learning-based methods, we achieve higher accuracy than Meta-Learner both in the 1-shot and the 5-shot set-ups, while being just slightly behind MAML. Compared with metric-based methods, which are more similar to ours, we marginally outperform the MatchingNet.
% These results confirm our conjecture that low level visual cues within trained CNNs can naturally perform few-shot learning.

Moreover, we evaluate out methods' few-shot learning ability with variances of the settings (e.g., number of shots and number of categories) and the results are listed in Table~\ref{tab:flexible}. Here we use exactly the same model as the ones in Table~\ref{tab:miniimagenet} which are trained only once on the training and validation set. To our knowledge, all of the state-of-the-art few-shot learning methods listed in Table~\ref{tab:miniimagenet} cannot deal with changes in the number of categories easily. By contrast, our few-shot learning methods based on VC-Encoding can be extended with minimal re-training to any number of shots and any number of categories and consistently outperform baseline methods.

% We observe that on the 5-shot scenario, our likelihood model is significantly better than our nearest neighbor model. A possible explanation is that the likelihood model combines several training examples into a distribution while nearest neighbor can only use examples individually. For instance, if a front wheel of cars appears in one training example and a rare wheel occurs in another example, the likelihood model can combine these two wheels into a distribution while nearest neighbor can only match testing examples with either the front wheel or the rare wheel. 

\begin{figure}[t]
    \begin{center}
        \includegraphics[width=0.98\linewidth]{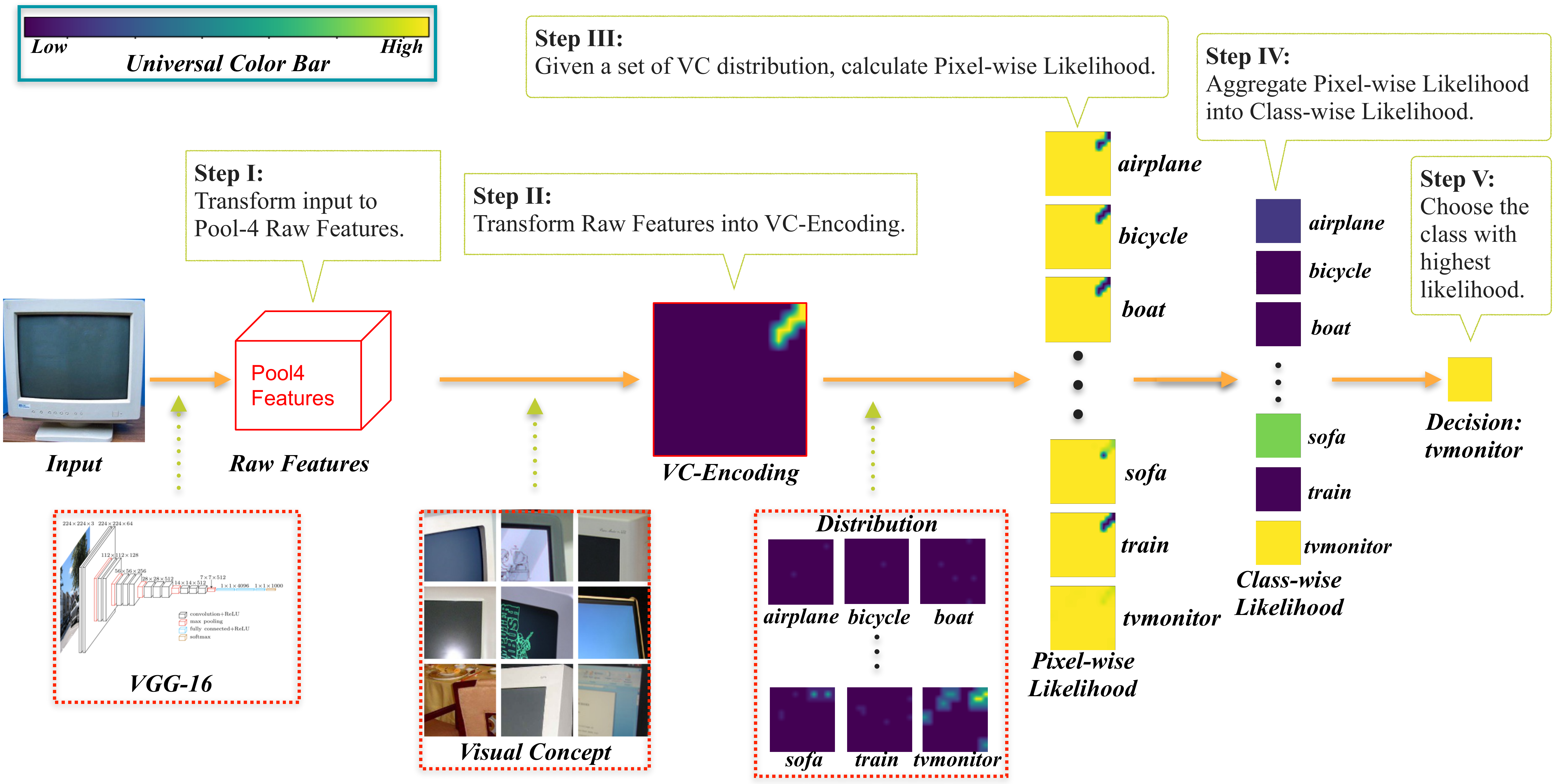}
    \end{center}
    \vspace{-10pt}
    \caption{Visualizing the inference procedure of the factorizable likelihood model (using one VC for example). For better visualization, we rescale the variance of aggregated likelihood to $1$. For all the visualizations, we use the same Universal Color Bar. This figure is best viewed in color. See Section \ref{subsec:pascal} for more details.}
\label{fig:procedure}
\end{figure}

\subsection{PASCAL3D+}
\label{subsec:pascal}
Here we apply our methods to PASCAL3D+~\cite{xiang2014beyond}, a dataset with larger high quality images than Mini-ImageNet. It was originally tailored for 3D object detection and pose estimation by augmenting $12$ rigid categories from PASCAL VOC 2012. Since we interpret our few-shot recognition mainly by visualizing every step of the inference, with input images of sufficient sizes, we can obtain large VC-Encoding distribution maps whose visualizations are easy for humans to understand.
%We choose PASCAL3D+ as our testbed since it provides high quality images with comparable image size to ImageNet.

% The simplicity of our methods makes the inference process of few-shot recognition very transparent. 
On PASCAL3D+, we first give intuition for the interpretability of our model. In Figure~\ref{fig:procedure}, we visualize every step of the inference process using our method based on an example VC. By looking at the closest patches for the VC of interest, we find this VC is very likely to relate to the corners of TV Monitors. Then, for a given test image, we convert its original VGG16 Pool-4 features into VC-Encoding and observe the example VC mainly appears at upper right corner of the encoding map. After calculating the pixel-wise likelihood using the distributions learned from a few training images in each candidate category, it is clearer that except for the TV Monitor, all other categories show low likelihood in the area of the corner. Finally, we aggregate the likelihood and make the correct classification decision by assigning the test image to the ``TV Monitor'' category.

Meanwhile, we quantitatively evaluate our methods on PASCAL3D+. To get a feature extractor CNN, we use a subset of the ImageNet~\cite{deng2009imagenet} classification dataset which excludes object categories that relate to the $12$ categories used in PASCAL3D+ ($956$ categories left), and we train an ordinary VGG-16 which achieves $71.27\%$ top-1 accuracy. For testing, we crop the objects out using annotated bounding boxes provided by PASCAL3D+ and resize them into $224\times224$. Then we use their Pool-4 features to implement our few-shot methods. As a comparison, we propose two baseline models. One is a nearest neighbor method based on raw Pool-4 features using the cosine distance metric; the other is an Exemplar-SVM trained using hinge loss. Both of them use the same pre-trained VGG-16 as our VC methods. During evaluation, we set $20$ trials of both 5-shot and 1-shot learning over $12$ categories on PASCAL3D+. We also assess our methods using different numbers of VCs. The results are shown in Table~\ref{tab:pascal}.

% In light of our testing results, we conclude that VC-Encoding is a useful semantic decomposition of images into parts.
In general, our VC-based methods consistently outperform two baselines by large margins. In particular, the improvements we achieve (especially in our nearest neighbor models) compared to the baseline methods are due to the use of VCs (e.g., by transferring the raw feature vectors into VC distances, and by thresholding the distance to get the VC-Encoding). Thus, we claim that decomposing fuzzy features (i.e., deep network features) into explicit semantic cues (i.e., the VCs) improves both interpretability and performance. We also notice that our methods are not sensitive to the number of VCs since changes of the number of VCs only cause slight differences among accuracies.

\begin{table}[t]\footnotesize
    \begin{center}
    \begin{tabular}{lccc} 
        \toprule
        \multirow{2}{*}{\textbf{Method}} & \multirow{2}{*}{\textbf{Number of VCs}} &
        \multicolumn{2}{c}{\textbf{12-category}} \\
        & & 1-shot & 5-shot \\
        \midrule
        \textbf{Pool4-NN}  & -- & \boldmath $36.12\%$ & $52.30\%$ \\ 
        \textbf{Pool4-SVM}  & -- & $32.66\%$ & \boldmath $52.46\%$ \\ 
        \midrule\midrule
        \textbf{VC-LH (Ours)} & $120$ & $39.25\%$ & $64.37\%$ \\
        \textbf{VC-LH (Ours)} & $200$ & \boldmath $40.02\%$ & $66.00\%$ \\
        \textbf{VC-LH (Ours)} & $300$ & $39.23\%$ & \boldmath $66.47\%$ \\
        \midrule
        \textbf{VC-NN (Ours)} & $120$ & $40.74\%$ & $58.52\%$ \\
        \textbf{VC-NN (Ours)} & $200$ & \boldmath $42.36\%$ & $59.47\%$ \\
        \textbf{VC-NN (Ours)} & $300$ & $41.18\%$ & \boldmath $61.07\%$ \\
        \bottomrule
    \end{tabular}
    \end{center}
    \vspace{-10pt}
    \caption{Average classification accuracies on PASCAL3D+. At the top are the baseline methods including nearest neighbor (Pool4-NN) and Exemplar-SVM (Pool4-SVM), based on Pool-4 features from the same VGG-16 used in our methods. In the middle and the bottom are our factorizable likelihood models (VC-LH) and VCs-based nearest neighbor models (VC-NN) using different number of VCs respectively. Marked in bold are the best results within each group for each scenario.}
\label{tab:pascal}
\end{table}

%% file: section6.tex
% !TEX root = main.tex

\section{Conclusion}
\label{sec:conclusion}

In this paper we address the challenge of developing simple interpretable models for few-shot learning by exploiting the internal representations of CNNs. We adapt the VCs from \cite{wang2015unsupervised} to the few-shot learning setting where the VCs are extracted from a small set of images of novel object categories using features from CNNs trained on other object categories. We extend the use of VCs by VC-Encoding and observe two properties, namely category sensitivity and spatial pattern, which leads us to propose two novel methods for few-shot learning that are simple and interpretable. Our methods show comparable performances and much superior flexibility to the current state-of-the-art methods -- it can be applied to a range of different few-shot scenarios with minimal re-training. In summary, we show that VCs and VC-Encodings enable ordinary CNNs to perform few-shot learning tasks. Future work may include improving the quality of the extracted VCs and extending our approach to few-shot detection.

% We emphasize that in this paper we have concentrated on developing the core ideas of our two few-shot learning models and that we have not explored variants of our ideas which could lead to better performance by exploiting standard performance enhancing tricks, or by specializing to specific few-shot challenges.